\documentclass[10pt,twocolumn,letterpaper]{article}

\usepackage{iccv}
\usepackage{times}
\usepackage{epsfig}
\usepackage{graphicx}
\usepackage{amsmath}
\usepackage{amssymb}
\usepackage{graphicx}
\usepackage{amsmath}
\usepackage{amssymb}
\usepackage{booktabs}
\usepackage{multirow}
\usepackage{colortbl}
\usepackage{color}
\usepackage{mathptmx}
\usepackage{makecell}

\usepackage[pagebackref=true,breaklinks=true,letterpaper=true,colorlinks,bookmarks=false]{hyperref}

\iccvfinalcopy 

\def\iccvPaperID{5237} 

\ificcvfinal\fi

\usepackage[capitalize]{cleveref}
\crefname{section}{Sec.}{Secs.}
\Crefname{section}{Section}{Sections}
\Crefname{table}{Table}{Tables}
\crefname{table}{Tab.}{Tabs.}

\definecolor{mygray}{gray}{.93}
\begin{document}

\title{SG-Former: Self-guided Transformer with Evolving Token Reallocation}

\author{
Sucheng Ren,~~~Xingyi Yang,~~~Songhua Liu,~~~Xinchao Wang\thanks{Corresponding Author: xinchao@nus.edu.sg} \\
National University of Singapore
}

\maketitle
\ificcvfinal\thispagestyle{empty}\fi

\begin{abstract}
Vision Transformer has demonstrated impressive success across various vision tasks. However, its heavy computation cost,
which grows quadratically with respect to the token sequence length, largely limits its power in handling large feature maps. 
To alleviate the computation cost, previous works rely on either fine-grained self-attentions restricted to local small regions, or global self-attentions but to shorten the sequence length resulting in coarse granularity. In this paper, we propose a novel model, termed as Self-guided Transformer~(SG-Former), towards effective global self-attention with adaptive fine granularity. At the heart of our approach is to utilize a significance map, which is estimated through hybrid-scale self-attention and evolves itself during training, to reallocate tokens
based on the significance of each region. Intuitively, we assign more tokens to the salient regions for achieving fine-grained attention,
while allocating fewer tokens to the minor regions in exchange for efficiency and global receptive fields. The proposed  SG-Former achieves performance superior to state of the art: our base size model achieves \textbf{84.7\%} Top-1 accuracy on ImageNet-1K, \textbf{51.2mAP} bbAP on CoCo, \textbf{52.7mIoU} on ADE20K surpassing the Swin Transformer by \textbf{+1.3\% / +2.7 mAP/ +3 mIoU}, 
with lower computation costs and fewer parameters. The code is available at \href{https://github.com/OliverRensu/SG-Former}{https://github.com/OliverRensu/SG-Former}
\end{abstract}

\section{Introduction}
Transformers~\cite{vaswani2017attention}, 
originated from natural language processing (NLP), 
have recently demonstrated state-of-the-art performance in
visual learning. 
The pioneering work of Vision Transformer (ViT)~\cite{dosovitskiy2020vit}
introduces the self-attention module
and explicitly models the long-range 
dependency between image patches,
which overcomes the inherent limitation of the local receptive field in convolution, thereby enhancing the performances of various 
tasks~\cite{cswin,swin,xie2021segformer,zheng2020rethinking,zhu2020deformable,carion2020end}.

\begin{figure}[t]
    \centering
    \begin{tabular}{c}
         \hspace{-5mm}\includegraphics[width=0.45\textwidth]{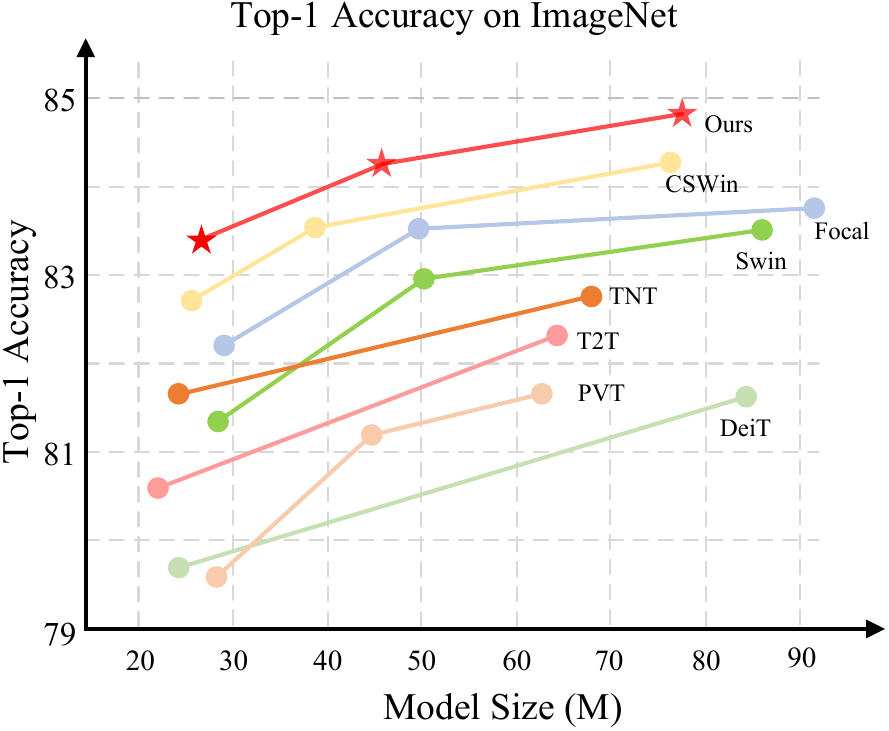}
    \end{tabular}
    \caption{Top-1 accuracy on ImageNet of recent Transformer models. Our proposed SG-Former yields significant improvements over all the baselines, including Swin Transformer and the state-of-the-art CSWin Transformer.}
  \label{fig:tesear2}
\end{figure}

\begin{figure*}[t]
    \centering
    \includegraphics[width=0.98\linewidth]{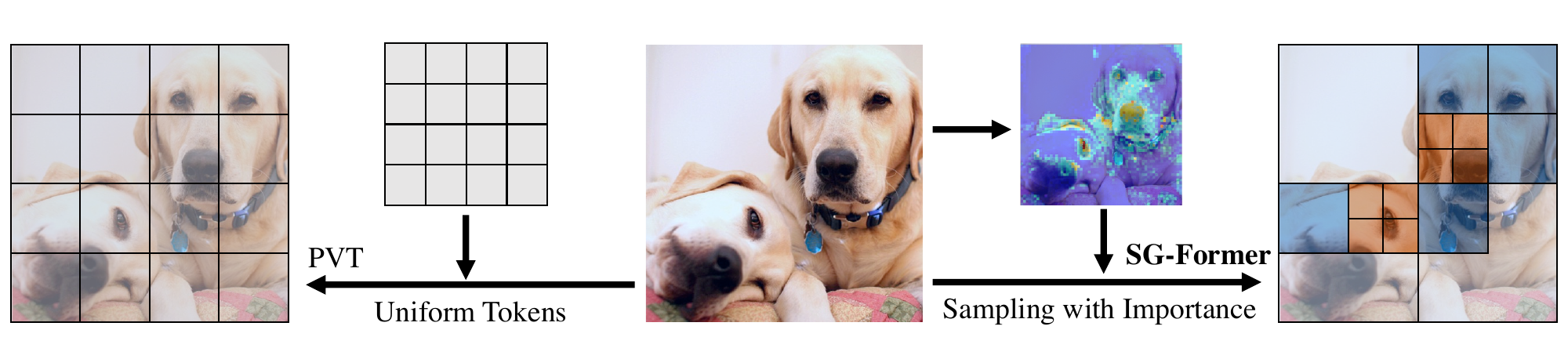}
    \caption{Visualization of our core idea. (Right) A significance map from the Transformer itself is used as guidance. Given an input sequence, SG-Former reallocates more tokens at salient regions (like the face of the dog) for fine-grained information and fewer tokens at backgrounds (like the wall) for the global receptive field with computation efficiency. (Left) PVT takes a predefined uniform map to aggregate tokens, regardless of their semantics.}
  \label{fig:tesear1}
\end{figure*}

Despite tremendous successes,  the computation cost of self-attention 
grows quadratically with respect to the sequence length,  which, in turn, greatly limits its applicability to large-scale inputs. 
To reduce the computation cost, ViT takes 
a large stride patch embedding
to reduce the sequence lengths. 
However, such operations inevitably 
result in self-attention applicable only to 
small-size feature maps with coarse granularity. 
To compute self-attention at high-resolution features, some approaches of~\cite{swin,cswin,vaswani2021scaling} are proposed
to restrict the self-attention region into a local window instead of the whole feature map (i.e., \emph{fine-grained local self-attention}). 
For example, Swin Transformer designs window attention, and CSWin designs cross-shape attention. As such, these methods~\cite{swin,cswin,vaswani2021scaling} sacrifice the power of modeling global information in each self-attention layer. Another stream of methods~\cite{pvt,focal,pvtv2} aims to aggregate tokens across the whole key-value feature maps to reduce the global sequence length (i.e., \emph{coarse-grained global attention}). For instance, Pyramid vision Transfomer (PVT)~\cite{pvt} uniformly aggregates tokens across the whole feature map with a large kernel of the large stride, which results in uniform coarse information over the whole feature map. 

In this paper, we introduce a novel Transformer model,
termed as self-guided Transformer~(SG-Former),
towards  \emph{global attention with adaptive fine granularity}
through an evolving self-attention design.
The core idea of SG-Former lies in that,
we preserve the 
long-range dependency across the whole feature map,
while reallocating tokens based on the 
significance of image regions.
Intuitively, we tend to assign more tokens
to the salient regions 
so that each token may interact
with salient regions at fine granularity, 
and meanwhile
allocate fewer tokens
over the minor regions
in exchange for efficiency. SG-Former keeps estimating self-attention efficiently with global receptive fields and attending to fine-grained information at salient regions adaptively. As shown in Figure \ref{fig:tesear1}, our SG-Former reallocates more tokens at salient regions like the dog and fewer tokens at minor regions like the wall according to the significance maps obtained from itself.  PVT, on the other hand, adopts predefined strategies to aggregate tokens uniformly. 

Specifically, we keep the query tokens but reallocate the key and value tokens for efficient global self-attention. The significance
of image regions, in the form of a score map,
is \emph{per se}
estimated through a hybrid-scale self-attention
and further utilized for guiding the 
token reallocation. 
In other words, given an input image,
the token reallocation is accomplished through
\emph{self guidance}, indicating that
each image undergoes a unique token reallocation
customized only for itself.
The reallocated tokens are, therefore, less influenced by humans prior.
In addition, such self-guidance
\emph{evolves} due to progressively accurate significance map prediction during training. The significance maps greatly affect the efficacy of the reallocation, and as such, we propose a hybrid-scale self-attention that accounts for various granularity information within one layer with the same cost as Swin. The various granularity information in hybrid-scale self-attention is achieved by grouping heads and diversifying each group for different attention granularity. The hybrid-scale self-attention also provides hybrid-scale information to the whole Transformer.

Our contributions are therefore summarized as follows.

\begin{itemize}
    \item We introduce a novel Transformer model, SG-Former,
    through unifying hybrid-scale information extraction, including fine-grained local and global coarse-grained information within one self-attention layer. With unified local global hybrid-scale information, we predict the significance map to identify regional significance.
    \item  With the significance map, we model self-guided attention to automatically locate salient regions and keep salient regions fine-grained for accurate extract information and minor regions coarse-grained for low computation cost.
    \item We evaluate our proposed Transformer backbone on various downstream tasks, including classification, object detection, and segmentation. Experimental results demonstrate that SG-Former consistently outperforms previous Vision Transformers under similar model sizes.
\end{itemize}

\section{Related Work}
\noindent\textbf{Vision Transformers.}
Vision Transformer (ViT)~\cite{vit} originated from Transformer\cite{vaswani2017attention} for sequences in natural language processing (NLP) first introduces a pure Transformer architecture into computer vision. Such architecture achieves impressive improvements in image classification over previous convolution nets. The requirements of large-scale datasets like JFT-300M in training ViT limits its application on various tasks. Some works~\cite{deit,Touvron2022DeiTIR,coadvise,tinymim,sdmp} propose complex data augmentation, regularization, knowledge distillation, and training strategies. Recently, some work design better Transformer architectures based on Vision Transformer for detection~\cite{wang2020end,sun2021rethinking}, segmentation~\cite{strudel2021segmenter,zheng2021rethinking,Ren_2021_CVPR}, restoration~\cite{Ren_2022_CVPR}, retrieval~\cite{el2021training}, re-id~\cite{he2021transreid}, with improved scale-ability~\cite{DBLP:conf/nips/RiquelmePMNJPKH21,DBLP:conf/iclr/ChenZJLW23} and robustness~\cite{li2023trade}. Different from these work, recent work~\cite{cswin,swin,focal,zhang2021multi,yuan2021tokens,touvron2021going,yuan2021incorporating,li2021localvit} aim at designing general Transformer backbone for general vision tasks instead of some specific tasks. The most significant difference between these Transformers and vanilla ViT is that these Transformers take hierarchical architecture following the same idea of convolution instead of the single-scale features in ViT and more efficient self-attentions in dealing with large-size feature maps. Notably, it has been demonstrated that the overall architecture~\cite{DBLP:conf/cvpr/YuLZSZWFY22,DBLP:journals/corr/abs-2210-13452}, rather than the specialized attention operation, plays a crucial role in achieving superior performance. As a result, this insight has gives rise to more powerful and efficient architectures~\cite{DBLP:journals/corr/abs-2305-12972,DBLP:conf/cvpr/0003MWFDX22}.
\begin{figure*}[t]
    \centering
    \begin{tabular}{c}
         \includegraphics[width=0.98\textwidth]{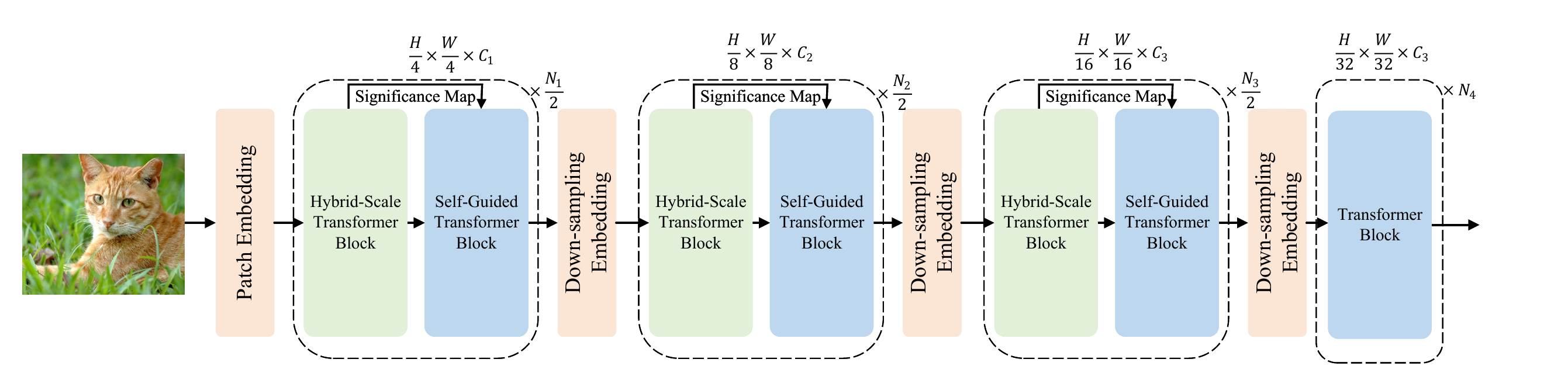}
    \end{tabular}
    \caption{The overall architecture of SG-Former. Hybrid-Scale Transformer blocks extract hybrid-scale information and provide significance for self-guided attention.}
  \label{fig:method}
\end{figure*}

\textbf{Efficient Deep Learning.} Designing efficient deep learning models with minimal training time~\cite{yang2022dery,DBLP:conf/icml/NoklandE19,TaskRes}, low inference cost~\cite{Hinton2015DistillingTK,DBLP:conf/iclr/ZagoruykoK17,DBLP:conf/eccv/YangYW22,fang2023depgraph,Xinyin2023structural,fang2023structural}, and reduced data requirements~\cite{wang2020dataset,DBLP:conf/nips/Liu0YYW22,slimmableDC} has been a long-standing pursuit. In the case of ViT, efficient self-attentions become crucial when dealing with a large number of tokens, such as processing long documents in NLP or high-resolution images in computer vision, where self-attention can incur quadratic computational and memory costs. On the one hand, local fine-grained self-attention limits the query token attending neighborhood tokens within a predefined window. Swin Transformer~\cite{swin} split feature maps into window shape regions to limit the self-attention locally. Axial attention~\cite{ho2019axial}
propose calculating stripe window attention. Focal~\cite{focal} Transformer predefined the coarse-fine granularity regions and attend different regions with different granularities. CSWin~\cite{cswin} designs windows with horizontal and vertical stripes. CSWin splits heads into parallel groups and applies different attention to different groups. On the other hand, global coarse-grained self-attention keeps the ability to model global dependency in self-attention but faces coarse-grained information processing. PVT~\cite{pvt} tries to keep the global receptive field with token reduction but leads to coarse-grained self-attention. Focal Transformer~\cite{focal} predefines more attention to close surroundings and less attention to far away positions. However, such inductive bias limits the model's ability to attend to salient objects far away from the current token position. Different from previous work, we propose to give the self-attention module more flexibility and let it find the best way to reduce the computation cost only with some predefined guidance instead of a specific method.


\section{Method}
\subsection{Overview}
The overall pipeline of SG-Former is shown in Figure \ref{fig:method}. SG-Former shares the same patch embedding layer and four-stage pyramid architecture as previous CNN and Transformer models~\cite{resnet,pvt,swin,cswin}. An image $X\in \mathbb{R}^{H\times W\times 3}$ is first downsampled $4 \times$ by the patch embedding layer at the input level. A downsampling layer with a rate $2\times$ exists between two stages. Therefore, the feature map is of size $\frac{H}{2^{i+1}}\times \frac{W}{2^{i+1}}$ at the $i$-th stage. Except for the last stage, each stage has $N_i$ Transformer 
blocks, consisting of two types of blocks repeatedly: (i) a hybrid-scale Transformer block and (ii) a self-guided Transformer block. The hybrid-scale self-attention extracts hybrid-scale objects and multi-granularity information, guiding regional significance; its detail will be covered in Section \ref{sec:hybrid-scale}. The self-guided self-attention models global information while keeping the fine-granularity at salient regions according to the significance information from the hybrid-scale Transformer block; its detail will be covered in Section \ref{sec:guidance}.

\subsection{Self-Guided Attention}
\label{sec:guidance}
\begin{figure}[t]
    \centering
    \begin{tabular}{c}
         \includegraphics[width=0.45\textwidth]{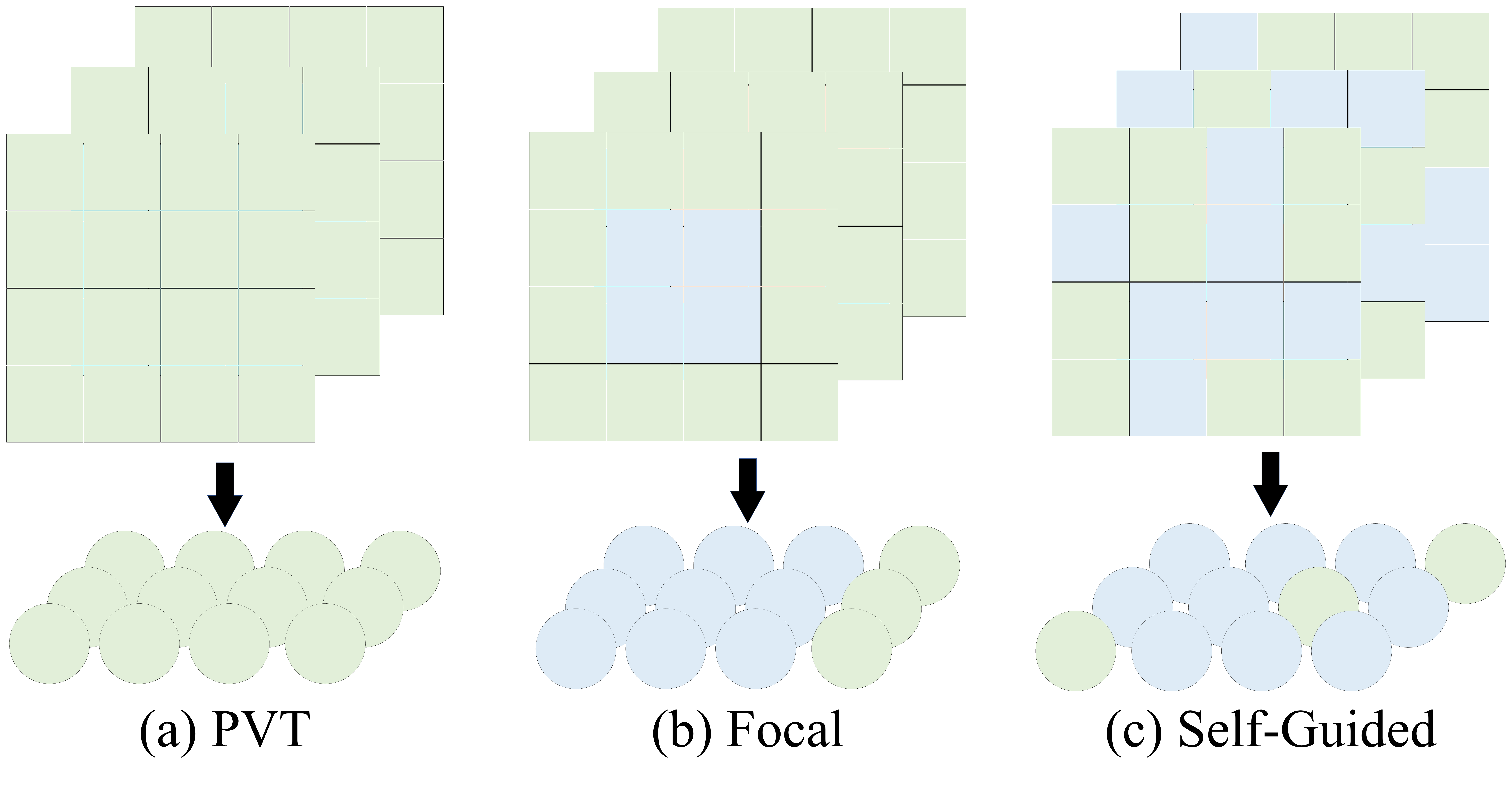}
    \end{tabular}
    \caption{Comparison of fusion. 
    (A) PVT treats all tokens/regions to be equal,
    indicating that their significance is uniform across the whole map. 
    All input images share the same aggregation strategies. 
    (B) Focal Transformer attempts to  preserve more tokens at central regions and fewer tokens at  background regions, indicating that 
    the significance is larger in the central regions and smaller at the background regions.
    All input images share the same aggregation strategies. 
    (C) Ours takes the model-predicted significance maps as guidance for aggregation, and preserves more tokens (the blue ones) in salient regions and fewer tokens in minor regions (the green ones). The significance depends on the input images and the model.}
  \label{fig:fusion}
\end{figure}
Despite the capability of modeling extended range information, the high computation cost and memory consumption of vanilla self-attention are quadratic versus the sequence length, limiting its applicability on large-size feature maps in various computer vision tasks, such as segmentation and detection. Recent works~\cite{pvt,pvtv2} suggest reducing sequence length with token aggregation by merging a few tokens into one. However, such aggregation treats each token equally and ignores the inherent significance difference of different tokens. Such aggregation faces two problems: (i) information might lose or mix with irrelevant information at salient regions, and (ii) at a minor region or background region, a lot of tokens (a higher proportion of sequences) are redundant for simple semantics while demanding a lot of computation. Motivated by this observation, we propose self-guided attention, which uses significance as guidance to aggregate tokens. Namely, in salient regions, more tokens are preserved for fine-grained information, while in minor regions, fewer tokens are maintained to keep the global view of self-attention and reduce computation cost at the same time. As shown in Figure \ref{fig:fusion}, "self-guided" indicates that the Transformer itself determines our computation cost reduction strategy during its training, instead of human-introduced prior knowledge, such as window attentions in Swin~\cite{swin}, cross-shape attention in CSWin~\cite{cswin}, static spatial reduction in PVT~\cite{pvt}.

The input feature map $X \in \mathbb{R}^{h\times w \times c}$ are firstly projected to Query($Q$), Key$(K)$ and Value$(V)$. Next, $H$ independent self-attention heads compute self-attention in parallel. To reduce the computation cost while keeping the size of feature maps unchanged after attention, we fix the length of Q but aggregate tokens of $K$ and $V$ with \textbf{i}mportance guided \textbf{a}ggregation \textbf{m}odule (IAM). 
\begin{equation}\label{eq:iam}
\begin{aligned}
    Q &= XW^Q, \\
    K&= \mathrm{IAM}(X, S, r)W^K, \\
    V &= \mathrm{IAM}(X, S, r)W^V.
\end{aligned}
\end{equation}
The goal of IAM is aggregate fewer tokens to one (i.e., preserve more) at salient regions and more tokens to one (i.e., preserve less) at background regions. In Eq. (\ref{eq:iam}), the significance map $S \in \mathbb{R}^{h \times w}$ contains the information of regional significance with multiple granularities (see Section~\ref{sec:hybrid-scale}). We sort the value of the significance map in ascending order and evenly divide S into $n$ sub-regions $S^1, \cdots, S^n$. Consequently, $S^n$ and $S^1$ are the most important and minor regions, respectively. Meanwhile, all tokens in $X$ are grouped into $X^1, \cdots,X^n$ according to $S^1, \cdots,S^n$. In Eq. (\ref{eq:iam}), $r$ represents the aggregation rate that every $r$ tokens are aggregated into one token. We set different aggregation rate $r^1, \cdots, r^n$ at areas of different signficance, so that each sub-region has one aggregation rate, and the more important the sub region is, the smaller aggregation rate is. The specific values of $r$ for different stages are listed in Table \ref{tab:variant}. Therefore, the IAM aggregate grouped input features $X^1, \cdots,X^n$ group by group  each group with different aggregation rate reallocated tokens by concatenating them of each group.
\begin{equation}
\begin{split}
    &\hat{X}^i = F(X^i, r_i),\\
    &\hat{X} = \mathrm{Cat}(\hat{X}^1,\cdots,\hat{X}^n) 
\end{split}
\end{equation}
where $F(X, r)$ is the aggregation function, we implement it by fully connected layers with input dimension of $r$ and output dimension of 1. The number of tokens in $\hat{X}_i$ equals the number of tokens in $X^i$ divided by $r_i$.

\subsection{Hybrid-scale Attention}
\label{sec:hybrid-scale}
\begin{figure}[t]
    \centering
    \begin{tabular}{c}
         \includegraphics[width=0.48\textwidth]{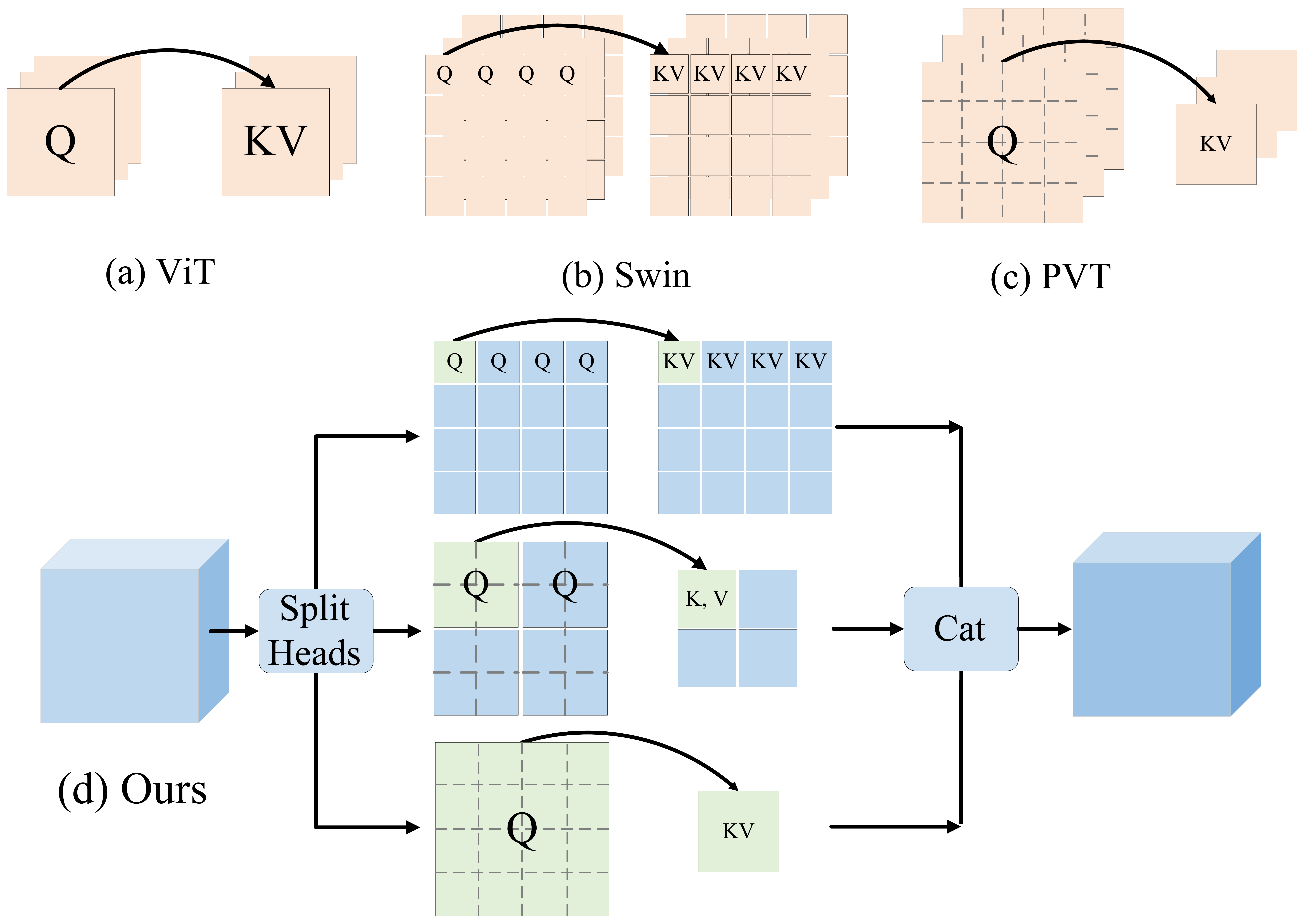}
    \end{tabular}
    \caption{Comparison of Local Global information and hybrid-scale. (A) Self-attention in ViT is single scale and has huge computation cost, 
    thereby applicable to only small  feature maps. (B) Self-attention in Swin Transformer is single scale and is limited within small windows, which reduces computation cost and preserves fine-grained information; however, the attention is local. (C) Self-attention in PVT is single scale. It reduces tokens alongside computation cost, and preserves global receptive field; however, it captures coarse information only.  (D) Our Self-attention is hybrid-scale and performs attention from local to global areas. It  reduces computation cost, preserves global receptive field, and captures fine-grained hybrid-scale information.  }
  \label{fig:scale}
\end{figure}

\begin{table*}[t]
\centering
\begin{tabular}{c|c|c|c|c}
\toprule
                        & Size                      & SG-Former-S & SG-Former-M &SG-Former-B\\
\midrule
\multirow{3}{*}{Stage1} & \multirow{3}{*}{56x56}   &
$s$=1, 8&
$s$=1, 8&
$s$=1, 2, 4, 8\\
&&$r$=196, 56, 56, 28&$r$=196, 56, 56, 28&$r$=196, 56, 56, 28\\
&& $C_1$=64, $head$=2, $N_1$=2         &   $C_1$=64, $head$=2, $N_1$=2       &     $C_1$=96, $head$=4, $N_1$=4      \\
\midrule
\multirow{3}{*}{Stage2} & \multirow{3}{*}{28x28}   &
$s$=1, 4&
$s$=1, 4&
$s$=1, 2, 4\\
&&$r$=49, 14, 14, 7&$r$=49, 14, 14, 7&$r$=49, 14, 14, 7\\
&& $C_2$=128, $head$=4, $N_2$=4         &   $C_2$=128, $head$=4, $N_2$=6       &     $C_2$=192, $head$=6, $N_2$=6      \\
\midrule
\multirow{3}{*}{Stage3} & \multirow{3}{*}{14x14}   &
$s$=1, 2&
$s$=1, 2&
$s$=1, 2\\
&&$r$=2, 1&$r$=2, 1&$r$=2, 1\\
&& $C_3$=256, $head$=8, $N_4$=16         &   $C_3$=256, $head$=8, $N_4$=28       &     $C_3$=384, $head$=12, $N_4$=24     \\
\midrule
Stage4 & 7x7       &$C_4$=512, $head$=16, $N_4$=1          &   $C_4$=512, $head$=16, $N_4$=2       &    $C_4$=768, $head$=24, $N_4$=2  \\   
\bottomrule
\end{tabular}
\caption{Several model variants of our self-guided Transformer. $C$ and $N$ represent the dimension and number of blocks, respectively. $head$ indicates the number of heads. }
\label{tab:variant}
\end{table*}
The hybrid-scale attention serves for two purposes: (i) to extract hybrid-scale global and fine-grained information with no more computation cost than the window attentions in Swin Transformer, and (ii) to provide the significance for self-guided attention.

As shown in Figure \ref{fig:scale}, the input features $X$ are projected into Query(Q), Key(K), and Values(V). Then the multi-head self-attention adopts $H$ independent heads. Usually, these $H$ independent heads perform within the same local regions and thus lack head diversity. By contrast, we evenly dive $H$ heads into $h$ groups and inject hybrid-scale and multi-receptive field attentions into these $h$ groups, where there are $\frac{H}{h}$ heads in each group. In the $i$-th head which belong to the $j$-th group, with scale $s_j$ (where $j=1,\cdots,h$), every $s_j\times s_j$ tokens in $\{K, V\}$ are merged to one token. Next, we split $\{Q, K, V\}$ into windows. The window size of $\{K, V\}$ is set to $M$ and remains unchanged across all groups. To align the window size of $\{Q\}$ and $\{K, V\}$ with the token merge in $\{K, V\}$, window size of $\{Q\}$ is chosen to be $s_jM \times s_jM$, $s_j$ times larger than that of $\{K, V\}$. The receptive field of attention is significantly enlarged by $s_j$ times:
\begin{equation}
\begin{split}
    &Q_i = XW_i^Q,\\
    &K_i = \mathrm{Merge}(X, s_j)W_i^K,\\
    &V_i = \mathrm{Merge}(X, s_j)W_i^V,
\end{split}
\end{equation}
where $\mathrm{Merge}(X, s_j)$ indicates merging every $s_j\times s_j$ tokens in $X$ to one token, which is implemented with convolution of stride $s_j$. A special case is when $s_j$ equals $1$, where  no token merges and $\{Q, K, V\}$ have  windows of the same sizes. 

\begin{equation}
\centering
    \begin{split}
    Atten_i = \mathrm{Softmax}&\left(\frac{P(Q_i, s_jM)P(K_i^\mathsf{T}, M)}{\sqrt{d_h}}\right),\\
        h_i = &Atten_i P(V_i, M)
    \end{split}
\end{equation}
where $P(X, s_jM)$ indicates window partition with window size of $s_jM \times s_jM$. $Atten_j$ is the attention map. There is a special case: $s_jM\times s_jM$ equals to $h\times w$, 
where no window partition is required and all tokens in $\{K, V\}$ are attended by $\{Q\}$, leading to global information extraction.

The significance of the token is taken to be the sum of the products of all tokens and the current token:
\begin{equation}
    \begin{split}
        & S_i = \frac{1}{hw}\sum\limits_{m=1}^{hw} Atten_{i}^{m,n},\\
    & S = \sum\limits_{i=1}^{h} S_i,\\
    \end{split}
\end{equation}
where $S$ is the final significance map by summing over  $S_i$ for hybrid-scale guidance with both global and fine-grained information.


\subsection{Transformer Block}
With two self-attention mechanisms, we correspondingly design two types of Transformer blocks. These two Transformer blocks only differ at the attention layer, while all others are kept identical:

\begin{equation}
	\begin{split}
		\hat{X}^{i}  =  \mathrm{MS\text{-}Attention}(\mathrm{LN}(X^{i-1}))+X^{i-1}, \\
		\hat{X}^{i}  =  \mathrm{MLP}(\mathrm{LN}(X^{i-1}))+X^{i-1}, \\
	\end{split}
\end{equation}

\begin{equation}
	\begin{split}
		\hat{X}^{i}  =  \mathrm{SG\text{-}Attention}(\mathrm{LN}(X^{i-1}))+X^{i-1}, \\
		\hat{X}^{i}  =  \mathrm{MLP}(\mathrm{LN}(X^{i-1}))+X^{i-1}, \\
	\end{split}
\end{equation}
As shown in Figure~\ref{fig:method}, the first three stages are customized using our proposed hybrid-scale or self-guided Transformer blocks, while for the last stage, we use a vanilla Transformer block following previous Transformers~\cite{deit,vit,swin,cswin}. Note that the numbers of Transformer blocks at the first three stages (i.e., $N_1$, $N_2$, and $N_3$) are even, while that at the last stage (i.e., $N_4$) can be even or odd. 


\begin{table}[t]
\centering
\small
\begin{tabular}{l|c|c|c}
\toprule
\multirow{2}{*}{Model} & ~Params~ & ~~FLOPs~~ & ~~Top1~~ \\
                       & (M)     & (G) &     (\%) \\
\midrule
ResNet-50~\cite{resnet}    &       25.0     &     4.1      &    76.2      \\
RegNetY-4G~\cite{regnet} &       20.6     &  4.0         &     79.4     \\
EfficientNet-B4*~\cite{tan2019efficientnet} &       19    &  4.2        &     82.9     \\
DeiT-S~\cite{deit}         &      22.1      & 4.6          & 79.9        \\
DeepViT-S~\cite{zhou2021deepvit} & 27.0 &6.2 &82.3 \\
TNT-S~\cite{han2021transformer} &23.8  &5.2 & 81.3\\
CViT-15~\cite{chen2021crossvit}& 27.4 & 5.6&81.5 \\
PVT-S~\cite{pvt}       &      24.5      &     3.8      &   79.8      \\
Swin-T~\cite{swin} &      28.3      &   4.5        &    81.2     \\
Twin-S~\cite{twin} & 24 &2.9 &81.7 \\
LITv2-S~\cite{panfast} &28&3.7&82.0 \\
Focal-T~\cite{focal}  &        29.1    &    4.9       &    82.2     \\
ResTv2-T~\cite{zhang2022rest}& 30& 4.1&82.3 \\
PerViT-S~\cite{minperipheral}&21 &4.4& 82.1 \\
CSWin-T~\cite{cswin} &    23      &      4.3       &    82.7      \\ 
CETNet-T~\cite{wang2022convolutional}&23& 4.3& 82.7 \\
\rowcolor{mygray}
SG-Former-S &        22.5   &      4.8     &     \textbf{83.2} \\
\midrule
ResNet-101~\cite{resnet}       &     45.0     &    7.9     &  77.4     \\
EfficientNet-B5*~\cite{tan2019efficientnet} &       30    &  9.9        &     83.6     \\
Swin-S~\cite{swin} &     49.6    &   8.7   &  83.1     \\
PVT-M~\cite{pvt}  &     44.2   &    8.7  &   81.2  \\
Twin-B~\cite{twin} & 56 &8.6 &83.2 \\
Focal-S~\cite{focal} &     51.1   &   9.1   &  83.5    \\
CSWin-S~\cite{cswin} &      35   &   6.9 &  83.6   \\
\rowcolor{mygray}
SG-Former-M &        38.7  &     7.5     &     \textbf{84.1} \\
\midrule
EfficientNet-B6*~\cite{tan2019efficientnet} &       43    &  19.0        &     84.0     \\
ViT-B~\cite{vit} &     86.6     &    17.6     &   77.9     \\
DeiT-B~\cite{deit}   &     86.6     &     17.5     &  81.8    \\
Swin-B~\cite{swin} &      87.8    &   15.4   &   83.4     \\
Twin-L~\cite{twin} & 99.2 &15.1 &83.7 \\
PVT-L~\cite{pvt}  &     61.4    &    9.8  &   81.7   \\
LITv2-B~\cite{panfast} &87& 13.2& 83.6 \\
CETNet-B~\cite{wang2022convolutional}& 75& 15.1& 83.8 \\
Focal-B~\cite{focal} &      89.8    &    16.0   &  83.8    \\
CSWin-B~\cite{cswin} &      78   &    15.0  &  84.2   \\
ResTv2-L~\cite{zhang2022rest}& 87& 13.8&84.2 \\
\rowcolor{mygray}
SG-Former-B       &     77.9   &   15.6      &   \textbf{84.7}      \\
\bottomrule
\end{tabular}
\caption{Comparison of different backbones on ImageNet-1K classification. Except EfficientNet (EffNet-B4*), all models are trained and evaluated on the input size of $224\times 224$. Our  Transformer consistently outperforms previous state-of-the-art ones under all small, medium, and base model sizes.}
\label{classification}
\end{table}

\subsection{Transformer Architecture Variants}
We build three models of small (SG-Former-S), medium (SG-Former-M), and base (SG-Former-B) sizes to fairly compare with other Transformers. Their differences including channels, heads, and number of blocks in each stage are detailed in Table \ref{tab:variant}. Both small and medium models have two scales in each stage, while the base model has four scales at the first stage and three scales in the second stage. Following~\cite{cswin,shunted}, we also adopt convolution patch embedding, locally-enhanced detail-specific feedforward layer, and for simplicity, these details are not shown in equations.

\section{Experiments}

\begin{table}[t]
    \centering
    \begin{tabular}{c|c|c|c}
    \toprule
         \multirow{2}{*}{Model} & ~Params~ & ~~FLOPs~~ & ~~Top1~~ \\
                       & (M)     & (G) &     (\%) \\
\midrule
CvT-13& 20&16.3&83.0\\
T2T-14& 22&17.1&83.3\\
CViT-15& 28&21.4&83.5\\
CSWin-T& 23&14.0&84.3\\
\rowcolor{mygray}
         SG-Former-S& 23&15.1&84.7 \\
         \midrule
         CvT-21& 32&24.9&83.3\\
         CViT-18& 45&32.4&83.9\\
         CSWin-T& 35&22.0&85.0\\
         \rowcolor{mygray}
         SG-Former-M& 39&23.4& 85.3\\
         \midrule
         ViT-B/15& 86&48.3&77.9\\
         DeiT-B& 86&55.4&83.1\\
         Swin-B& 88&47.0&84.2\\
         CSWin-B& 78&47.0&85.4\\
         \rowcolor{mygray}
         SG-Former-B& 78&47.9& 85.8\\
    \bottomrule
    \end{tabular}
    \caption{Classification results on ImageNet-1K with input resolution of 384$\times$384.}
    \label{tab:384}
\end{table}

To demonstrate the architectural superiority of SG-Former backbone, we conduct experiments on ImageNet-1K classification~\cite{deng2009imagenet}, COCO object detection~\cite{coco}, and ADE20K semantic segmentation~\cite{ade20k}. Besides, we ablate  the influence of single scale and hybrid-scale self-attention, as well as the influence of different guidance in self-evolving attention.

\subsection{Classification on ImageNet-1K}
To fairly compare with previous methods, we follow the same training  setting~\cite{deit,cswin}. Specifically, all models are trained for 300 epochs on ImageNet-1K with the input resolution of 224$\times$224 and batch size of 1024. For optimization, we take the AdamW optimizer with the weight decay of 0.05 and cosine learning rate decay with linear warm-up with 20 epochs, and the peak learning rate is set to 1e-3. For data augmentation and regularization, we keep the same as CSWin~\cite{cswin}, including RandAugment, Exponential Moving Average, data mixing, label smoothing, stochastic depth, and Random Erasing. During inference, we resize and center crop to 224$\times$224.

\begin{table*}[t]
\centering
{
\begin{tabular}{l|c|cccccc|cccccc}
\toprule
\multirow{2}{*}{Backbone} & Params &  \multicolumn{6}{c|}{Mask R-CNN 1$\times$ schedule} & \multicolumn{6}{c}{Mask R-CNN 3$\times$ schedule + MS} \\
                          &   (M)       &   $AP^b$    &  $AP^b_{50}$     &  $AP^b_{75}$     &   $AP^m$    &   $AP^m_{50}$    &  $AP^m_{75}$     &    $AP^b$    &  $AP^b_{50}$     &  $AP^b_{75}$     &   $AP^m$    &   $AP^m_{50}$    &  $AP^m_{75}$     \\
\midrule
Res50~\cite{resnet} &   44        &   38.0& 58.6 &41.4 &34.4& 55.1& 36.7 &41.0& 61.7& 44.9& 37.1& 58.4& 40.1 \\
PVT-S~\cite{pvt}    &    44       & 40.4& 62.9& 43.8& 37.8& 60.1& 40.3& 43.0& 65.3 &46.9& 39.9& 62.5& 42.8 \\
Swin-T~\cite{swin} & 48 & 42.2 &64.6& 46.2 &39.1 &61.6& 42.0 &46.0& 68.2 &50.2 &41.6& 65.1 &44.8\\
Twin-S~\cite{twin} & 44&  43.4 &66.0& 47.3 &40.3 &63.2& 43.4 &46.8& 69.2 &51.2 &42.6& 66.3 &45.8\\
Focal-T~\cite{focal} & 49 & 44.8 &67.7& 49.2& 41.0& 64.7& 44.2 & 47.2 &69.4 &51.9& 42.7& 66.5& 45.9 \\
CSWin-T~\cite{cswin}    & 42& 46.7& 68.6 &51.3& 42.2& 65.6& 45.4& 49.0 &70.7 &53.7& 43.6& 67.9& 46.6 \\
SG-Former-S & 41 & \textbf{47.4}& \textbf{69.0} &\textbf{52.0}& \textbf{42.6}& \textbf{65.9}& \textbf{46.0}& \textbf{49.6} &\textbf{71.1} &\textbf{54.5}& \textbf{44.0}& \textbf{68.3}& \textbf{46.9} \\
\midrule
Res101~\cite{resnet} &   63     &  40.4& 61.1 &44.2 &36.4& 57.7& 38.8 &42.8& 63.2& 47.1& 38.5& 60.1& 41.3 \\
PVT-M~\cite{pvt}  &    64       & 42.0& 64.4& 45.6& 39.0& 61.6& 42.1& 44.2& 66.0 &48.2& 40.5& 63.1& 43.5 \\
Swin-S~\cite{swin} & 69 & 44.8 &66.6& 48.9 &40.9 &63.4& 44.2 &48.5& 70.2 &53.5 &43.3& 67.3 &46.6\\
Twin-B~\cite{twin} & 76& 45.2 &67.6& 49.3 &41.5 &64.5& 44.8 &48.0& 69.5 &52.7 &43.0&66.8 &46.6\\
Focal-S~\cite{focal} & 71& 47.4 &69.8 & 51.9& 42.8& 66.6& 46.1 & 48.8 &70.5 &53.6& 43.8& 67.7& 47.2 \\
CSWin-S~\cite{cswin}& 54 & 47.9 & 70.1 & 52.6 &43.2 & \textbf{67.1} & 46.2
&50.0& 71.3 & 54.7 & 44.5 & 68.4 & 47.7 \\
SG-Former-M& 51& \textbf{48.2} &\textbf{70.3} & \textbf{53.1}& \textbf{43.6}& 66.9& \textbf{47.0} & \textbf{50.5} &\textbf{71.5} &\textbf{54.9}& \textbf{45.4}& \textbf{68.8}& \textbf{48.2} \\
\midrule
X101-64~\cite{xie2017resx}  & 101   & 42.8 & 63.8 & 47.3 & 38.4 & 60.6 & 41.3
& 44.4 & 64.9 & 48.8 & 39.7 & 61.9 & 42.6 \\
PVT-L\cite{wang2021pyramid}            & 81  & 42.9 & 65.0 & 46.6 & 39.5 & 61.9 & 42.5
& 44.5 & 66.0 & 48.3 & 40.7 & 63.4 & 43.7 \\
Twins-L~\cite{chu2021twins}   & 111 & 45.9 & ---- & ---- & 41.6 & ---- & ----
& ---- & ---- & ---- & ---- & ---- & ----\\
Swin-B~\cite{liu2021swin}              & 107 & 46.9 & ---- & ---- & 42.3 & ---- & ----
& 48.5 & 69.8 & 53.2 & 43.4 & 66.8 & 46.9\\
CSWin-B~\cite{cswin}   & 97  & 48.7 & 70.4 & 53.9 & 43.9 & 67.8 & 47.3& 50.8 & 72.1 & 55.8 & 44.9 &69.1& 48.3 \\
SG-Former-B& 95  & \textbf{49.2} & \textbf{70.6} & \textbf{54.3} & \textbf{68.2} & \textbf{68.1} & \textbf{47.7}& \textbf{51.3} & \textbf{72.4} & \textbf{56.0} & \textbf{45.2} & \textbf{69.6} & \textbf{48.8} \\
\bottomrule
\end{tabular}}
\caption{Object detection and instance segmentation with Mask R-CNN on COCO. Only 3$\times$ schedule has the multi-scale training. All backbones are pretrained on ImageNet-1K. }
\label{maskrcnn}
\end{table*}

In Table \ref{classification}, we compare SG-Former with state-of-the-art CNN and Transformer architectures. SG-Former consistently outperforms competitors under similar parameters and computation costs. Specifically, SG-Former achieves 83.2\%, 84.1\%, and 84.7\% Top-1 Accuracy under small, medium, and base models, respectively. Comparing the SG-Former backbone with the CNN backbone, the SG-Former backbone is the first Transformer backbone that outperforms the previous state-of-the-art CNN-based backbone EfficientNet, which has much larger input resolutions. Although EfficientNet comes from neural network architecture search, our manually designed SG-Former-S/M/B still makes +0.3/+0.5/+0.6 Top-1 accuracy improvements.

Compared with baseline Transformers DieT, our small-size model even significantly outperforms the base size model by +1.4 with only a quarter of parameters (86.6M$\rightarrow$ 24.2M) and computation cost(17.6G$\rightarrow$ 4.8G). SG-Former-S/M/B achieve +2.0/+1.0/+1.2 improvements over  Swin-T/S/B. Compared with state-of-the-art Transformer CSWin, SG-Former-S/M/B makes +0.5/+0.5/+0.4 improvements under similar parameters.

\noindent\textbf{Throughoutput and Performance.}
As shown in the right figure, we evaluate the throughputs of different models on a single RTX2080Ti. Our SG-Former is the \emph{fastest} under similar performance or has the \emph{best} performance under similar inference speed.
\begin{figure}[t]
    \centering
    \includegraphics[width=.9\linewidth]{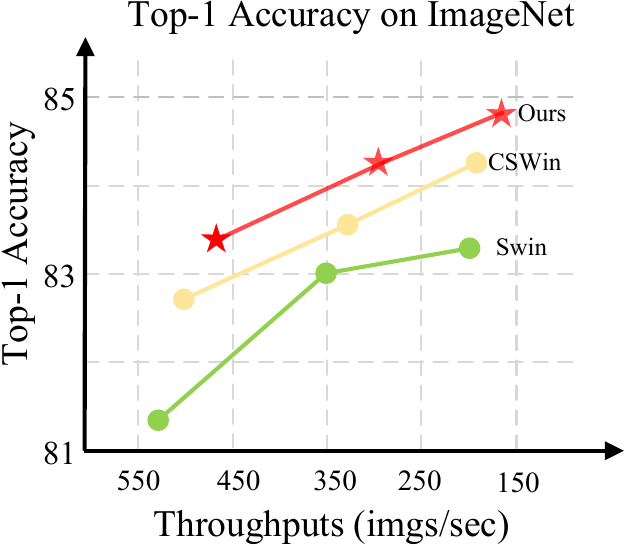}
    \caption{Top-1 accuracy  on ImageNet and throughputs of recent Transformer models. }
    \label{speed}
\end{figure}

\noindent\textbf{Classification with Higher Input Resolution.}
Following the same settings of \cite{cswin,swin}, we finetune our model with input resolution of 384$\times$ 384. With pretrained model on ImageNet-1K with 224$\times$224 input resolution, we finetune it for 30 epochs. We take AdamW as optimizer with weight decay of 1e-8. The learning rate is set to 5e-6 and the stochastic drop path is set to 0.4.

As shown in Table \ref{tab:384}, SG-Former significantly outperforms its counterparts under similar parameters. Specifically, our base size model outperforms Swin-B by \textbf{1.6}. SG-Former-S/M/B outperform previous state-of-the-art CSWin by \textbf{{+0.4, +0.3, +0.4}} respectively.

\subsection{Object Detection and Instance Segmentation}

\begin{table*}[t]
\centering
\begin{tabular}[t]{l|ccc|cccc}
\toprule
\multirow{2}{*}{Backbone} & \multicolumn{3}{c|}{Semantic FPN 80k} & \multicolumn{4}{c}{Upernet 160k}\\
 &   \#Param. & FLOPs & mIoU  &   \#Param. & FLOPs & mIoU&MS mIoU \\ 
\midrule
Res50~\cite{he2016deep}             & 28.5 & 183 & 36.7 & ----  & ---- & ----&---- \\
PVT-S~\cite{wang2021pyramid}        & 28.2 & 161 & 39.8 & ----  & ---- & ----&---- \\
Twins-S~\cite{chu2021twins}         & 28.3 & 144 & 43.2 & 54.4  & 901  & 46.2&47.1 \\
Swin-T~\cite{liu2021swin}           & 31.9 & 182 & 41.5 & 59.9  & 945  & 44.5&45.8 \\
Focal-T~\cite{focal} & - & - & - &62.0& 998& 45.8& 47.0 \\
CSWin-T                      & 26.1 & 202 & 48.2 & 59.9  & 959  & 49.3&50.7 \\
SG-Former-S                      & 25.4 & 205 &  \textbf{49.0} & 52.5  & 989  & \textbf{49.9}&  \textbf{51.5} \\
\midrule
Res101~\cite{he2016deep}            & 47.5 & 260 & 38.8 & 86.0  & 1029 & ----&44.9 \\
PVT-M~\cite{wang2021pyramid}        & 48.0 & 219 & 41.6 & ----  & ---- & ----&---- \\
Twins-B~\cite{chu2021twins}         & 60.4 & 261 & 45.3 & 88.5  & 1020 & 47.7&48.9 \\
Swin-S~\cite{liu2021swin}           & 53.2 & 274 & 45.2 & 81.3  & 1038 & 47.6&49.5 \\
Focal-S~\cite{focal}           & -& -& - & 85.0& 1130& 48.0 &50.0 \\
CSWin-S                      & 38.5 & 271 & 49.2 & 64.6  & 1027 & 50.4&51.5 \\
SG-Former-M            & 38.2 & 273 & \textbf{50.1} & 68.3 & 1114 & \textbf{51.2}&\textbf{52.1} \\
\midrule
X101-64~\cite{xie2017resx}          & 86.4 & --- & 40.2 & ----  & ---- & ----&---- \\
PVT-L~\cite{wang2021pyramid}        & 65.1 & 283 & 42.1 & ----  & ---- & ----&---- \\
Twins-L~\cite{chu2021twins}         & 103.7& 404 & 46.7 & 133.0 & 1164 & 48.8&50.2 \\
Swin-B~\cite{liu2021swin}           & 91.2 & 422 & 46.0 & 121.0 & 1188 & 48.1&49.7 \\
Focal-B~\cite{focal}  &- &- &- & 126.3 & 1354 &49.0  &50.5 \\
CSWin-B                      & 81.2 & 464 & 49.9 & 109.2 & 1222 &51.1&52.2\\
SG-Former-B                      & 81.0 & 475 & \textbf{50.6} & 109.3 & 1304 & \textbf{52.0}&\textbf{52.7}\\
\bottomrule
\end{tabular}
\caption{Performance comparison of different backbones on the ADE20K segmentation task. Two different frameworks, semantic FPN and Upernet, are used. }
\label{tab:segmentation}
\end{table*}

We evaluate the SG-Former backbone on object detection and instance segmentation with ImageNet pretraining as initialization and Mask R-CNN~\cite{he2017mask} as detection pipelines. All models are trained and evaluated on COCO2017~\cite{coco} with 118k images for training and 5K images for validation. We follow the same settings~\cite{swin,pvt,cswin} to use 1$\times$ schedule with 12 epoch training and 3$\times$ with 36 epochs training. For 1$\times$ schedule, we take single scale training with the input images of the shorter side to 800 while keeping the longer side no more than 1333. For 3 $\times$ schedule, we use multi-scale training by randomly resizing its shorter side to the range of [480, 800]. We compare SG-Former backbone with CNN backbones: ResNet~\cite{resnet}, ResNeXt(X)~\cite{xie2017aggregated} and Transformer backbone Swin\cite{swin}, PVT\cite{pvt}, Focal\cite{focal}, CSWin\cite{cswin}, Twin\cite{twin}.

In table \ref{maskrcnn}, we report the results with the MaskRCNN~\cite{he2017mask} pipeline and take bounding box mAP and mask mAP as an evaluation metric for object detection and instance segmentation, respectively. Compared with the CNN backbone, our method makes incredible improvements over the ResNeXt by +6.4 with 1$\times$ schedule and by +6.9 with 3$\times$ schedule. Compared with the Transformer backbones, SG-Former also shows superiority. SG-Former improves over Swin Transformer (the most well-known Transformer) and CSWin Transformer (the state-of-the-art one). Such supremacy comes from the ability to preserve more fine-granularity information and the global receptive field in our model.

\subsection{Semantic Segmentation on ADE20K}

We further evaluate SG-Former backbone on semantic segmentation with ADE20K dataset. We take semantic FPN, UperNet as basic framework. ADE20K is a widely-used semantic segmentation dataset with 150 semantic categories and contains 20K images for training and 2K images for validation. Following the same settings~\cite{cswin}, we take Semantic FPN and Upernet as the framework and implement them by mmsegmentation~\cite{mmseg2020}. For SemanticFPN, we take AdamW with weight decay of 0.0001 as the optimizer and the learning rate to be also 0.0001 for 80K iterations. We take mIoU as evaluation metric. For UperNet, we take AdamW with weight decay of 0.01 as the optimizer for 160K iterations with the batch size of 16. The learning rate is 6 $\times$ 10e-5 with 1500 iteration warmup at the beginning of training and linear learning rate decay. The augmentations include random flipping, random scaling and random photo-metric distortion. The input size is 512 × 512 in training. For SemanticFPN, we take AdamW with weight decay of 0.0001 as the optimizer and the learning rate to be also 0.0001 for 80K iterations. We take mIoU as the evaluation metric, and we report single-scale test results of Semantic FPN, both single-scale and multi-scale (MS) testing results of UperNet.

The results are reported in Table \ref{tab:segmentation}, SG-Former outperforms all competitors under similar parameters. Specifically, when using Semantic FPN, SG-Former achieves 50.6 mIoU and outperforms Swin Transformer by 4.6 mIoU with 10\% model size down. When the framework is UperNet, our SG-Former achieves 52.7 mIoU and outperforms Swin by 3 mIoU.

\subsection{Ablation Study}
\textbf{Single Scale and Hybrid Scale Self-Attention.}
Our hybrid-scale Transformer blocks have two functions: 1) capturing and identifying objects of different sizes with hybrid-granularity information. 2) Proving hybrid-scale significance to self-guided attention for aggregating background tokens and preserving more fine-granularity information at salient regions. 

We conduct ablation studies on the function of capturing and identifying objects of different sizes with hybrid-granularity information in Table \ref{tab:scale}. Single Scale (Local) indicates there is no token aggregation but just window attention and there is only one scale, while Single Scale (Global) indicates the aggregation rate is large enough $sM\times sM=H\times W$, the self-attention performs globally, and there is also only one scale. Our hybrid-scale self-attention outperforms local and global attention by $\textbf{1.0}$ and $\textbf{0.6}$ respectively

\begin{table}[]
	\centering
	\begin{tabular}{l|c|c|c}
	\toprule
		Model & Param. (M) & Flops (G) & Top-1 (\%) \\
		\hline
		Single Scale     &  \multirow{2}{*}{25.6}      &   \multirow{2}{*}{4.9}     &   \multirow{2}{*}{81.6}   \\
   (Local) &&& \\
   \hline
		Single Scale      &  \multirow{2}{*}{28.4}    &   \multirow{2}{*}{ 4.6}    &   \multirow{2}{*}{82.0}   \\
   (Global) &&&\\
  \hline
		Hybrid-Scale     &    \multirow{2}{*}{26.5}    &   \multirow{2}{*}{4.8}     &   \multirow{2}{*}{ 82.6}\\
   (Local-Global) &&&\\
		\bottomrule
	\end{tabular}
	\caption{Comparison of different scales in hybrid-scale self-attention.}
	\label{tab:scale}
\end{table}

\begin{table}[]
	\centering
	\begin{tabular}{l|c|c|c}
\toprule
Guidance & Param. (M) & Flops (G) & Top-1 (\%) \\
\midrule
Manual defined& 21.6& 4.6&82.3\\
Local&20.9&4.9& 82.9 \\
Global&24.6&4.6&82.4\\
Hybrid-Scale&22.5& 4.8& 83.2\\
\bottomrule
\end{tabular}
	\caption{Comparison of different significance maps. Manual defined indicate the significance equals at all position cross the whole significance map.}
	\label{tag:score}
\end{table}

\textbf{Significance Guidance}
To evaluate the influence of the significance map, in each module, we keep the first hybrid-scale Transformer block the same but change the source of important score maps in the second self-guided Transformer block. We manually define the significance map 
to be equal at all positions across the whole significance map as a comparison. The hybrid-scale self-attention layer provides significance maps with hybrid-scale information, yet we  use significance from the global scale or the local scale only.

As shown in Table \ref{tag:score}, the hybrid-scale significance outperforms the manually defined baseline which regards all region importance equally by \textbf{+0.9}. The local-scale significance which contains local fine-grained region importance information has better performance than that of the global scale which contains global coarse-grained region importance information, while hybrid-scale significance maps show much better performance over all competitors.

\section{Conclusion}
In this paper, we propose a novel Transformer backbone,
termed as Self-guided Transformer~(SG-Former). The key motivation is to aggregate tokens according to the region or token significance with hybrid-scale guidance, which explicitly
accounts for 
long-range dependency crosses the whole feature maps and attends more foreground tokens for fine-granularity information while merging more background tokens for computation efficiency. Experiments on classification, detection, and segmentation demonstrate 
that, the proposed 
SG-Former yields gratifying results:
our base size model achieves
a $84.7\%$ Top-1 accuracy on ImageNet-1K, 51.2mAP bbAP on
CoCo, 52.7mIoU on ADE20K, 
surpassing the Swin Transformer by {+1.3\% / +2.7 mAP/ +3 mIoU}, yet with a lower computation cost.

\section*{Acknowledgement}
This research is supported by the National Research Foundation, Singapore under its AI Singapore Programme (AISG Award No: AISG2-RP-2021-023).

{\small
\bibliographystyle{ieee_fullname}
\bibliography{egbib}
}

\end{document}